\def\year{2020}\relax
\documentclass[letterpaper]{article} 
\usepackage{aaai20}  
\usepackage{times}  
\usepackage{helvet} 
\usepackage{courier}  
\usepackage[hyphens]{url}  
\usepackage{graphicx} 
\usepackage{graphicx}  
\usepackage{todonotes}
\usepackage{xcolor}
\usepackage{amssymb}
\usepackage{amsmath}
\usepackage{bbm}
\usepackage{array}
\title{MA-DST: Multi-Attention-Based Scalable Dialog State Tracking}

\author{ \Large \textbf{Adarsh Kumar\textsuperscript{\rm 1}\thanks{Work done during internship at Amazon Alexa AI}, Peter Ku\textsuperscript{\rm 2}, Anuj Goyal\textsuperscript{\rm 2}, Angeliki Metallinou\textsuperscript{\rm 2}, Dilek Hakkani-Tur\textsuperscript{\rm 2}}\\ 
\textsuperscript{\rm 1}University of Wisconsin-Madison, \textsuperscript{\rm 2}Amazon Alexa AI, Sunnyvale, CA, USA\\ 
adarsh@cs.wisc.edu, \{kupeter, anujgoya, ametalli, hakkanit\}@amazon.com\\
}

\begin{document}

\maketitle

\begin{abstract}
Task oriented dialog agents provide a natural language interface for users to complete their goal. Dialog State Tracking (DST), which is often a core component of these systems, tracks the system's understanding of the user's goal throughout the conversation. To enable accurate multi-domain DST, the model needs to encode dependencies between past utterances and slot semantics and understand the dialog context, including long-range cross-domain references. We introduce a novel architecture for this task to encode the conversation history and slot semantics more robustly by using attention mechanisms at multiple granularities. In particular, we use cross-attention to model relationships between the context and slots at different semantic levels and self-attention to resolve cross-domain coreferences. In addition, our proposed architecture does not rely on knowing the domain ontologies beforehand and can also be used in a zero-shot setting for new domains or unseen slot values. Our model improves the joint goal accuracy by 5\% (absolute) in the full-data setting and by up to 2\% (absolute) in the zero-shot setting over the present state-of-the-art on the MultiWoZ 2.1 dataset. 
\end{abstract}

\section{Introduction}
\begin{figure}
	\centering
	\includegraphics[width=8cm,height=6cm]{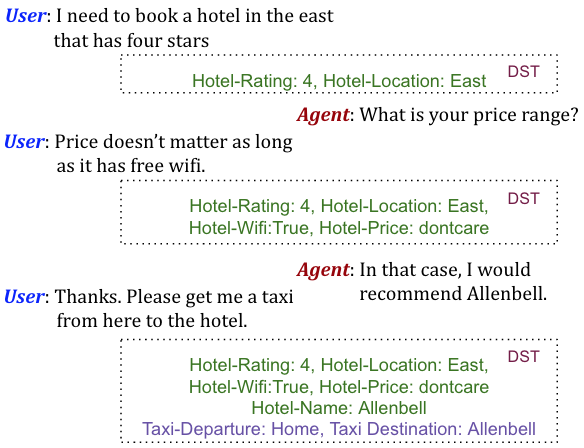}
	\caption{Sample multi-domain dialog, spanning hotel and taxi domains, along with its dialog state}
	\label{fig:example_dialog}
\end{figure}

Task-oriented dialog systems provide users with a natural language interface to
achieve a goal. Modern dialog systems support complex goals that may span
multiple domains. For example, during the dialog the user may ask for a hotel
reservation (hotel domain) and also a taxi ride to the hotel (taxi domain), as
illustrated in the example of Figure \ref{fig:example_dialog}. Dialog state
tracking is one of the core components of task-oriented dialog systems. The
dialog state can be thought as the system's belief of user's goal given the
conversation history. For each user turn, the dialog state commonly includes the
set of slot-value pairs, for all the slots which are mentioned by the user.
An example is shown in Figure \ref{fig:example_dialog}. Accurate DST is critical
for task-oriented dialog as most dialog systems rely on such a state to predict
the optimal next system action, such as a database query or a natural
language generation (NLG) response.


Dialog state tracking requires understanding the semantics of the agent and user
dialog so far, a challenging task since a dialog may span multiple domains and
may include user or system references to slots happening earlier in the
dialog. Data scarcity is an additional challenge, because dialog data collection
is a costly and time consuming \cite{kang2018data,lasecki2013conversations}. As
a result, it is critical to be able to train DST systems for new domains with
zero or little data.

Previous work formulates DST as a classification task over all possible slot
values for each slot, assuming all values are available in advance (e.g. through
a pre-defined
ontology) \cite{mrksic-etal-2017-neural,gao2019dialog,liuLane2017}. However, DST systems should be able to track the values of even free-form
slots such as $``hotel-name''$, which typically contain out-of-vocabulary
words. To overcome the limitations of ontology-based approaches candidate-set
generation based approaches have been proposed~\cite{46399}. 
TRADE~\cite{trade2019} extends this idea further and propose a decoder-based
approach that uses both generation and a pointing mechanism, taking a weighted
sum of a distribution over the vocabulary and a distribution over the words in
the conversation history. This enables the model to produce unseen slot values,
and it achieves state-of-the art results on the MultiWOZ public
benchmark \cite{budzianowski-etal-2018-multiwoz,ericMultiWOZ21}.
          
We extend this work by \cite{trade2019} and focus on improving the
encoding of dialog context and slot semantics for DST to robustly capture
important dependencies between slots and the conversation history as well as
long-range coreferences in the conversation history. For this purpose, we propose a Multi-Attention DST
(MA-DST) network. It contains multiple layers of cross-attention between the
slot encodings and the conversation history to capture relationships at
different levels of granularity, which are then followed by a self-attention
layer to help resolve references to earlier slot mentions in the dialog. We show
that the proposed MA-DST leads to an absolute improvement of over 5\%  in the
joint goal accuracy over the current state-of-the art for the MultiWOZ 2.1
dataset in the full-data setting. We also show that MA-DST can be adapted to new
domains with no training data in that new domain, achieving upto a 2\% absolute
joint goal accuracy gains in the zero-shot setting.


%

\section{Related Work}
Dialog state tracking (DST) is a core dialog systems problem that is well studied in the literature. Earlier approaches for DST relied on Markov Decision Processes (MDPs)~\cite{levin2000stochastic} and partially observable MDPs (POMDPs)~\cite{williams2007partially,Thomson:2010} for estimating the state updates. See \cite{williams2016the} for a review of DST challenges and earlier related work.


Recent neural state tracking approaches achieve
state-of-the-art performance on DST \cite{gao2018neural}. Some
of this work formulates the state tracking problem as a classification task over
all possible slot-values per
slot \cite{mrksic-etal-2017-neural,wen-etal-2017-network,liuLane2017}. This
assumes that an ontology containing all slot values per slot is available in
advance. In practice, this is a limiting assumption, especially for free-form
slots that may contain values not seen during training \cite{xu-hu-2018-end}. To

address this limitation, \cite{46399} propose a candidate generation approach
based on a bi-GRU network, that selects and scores slot values from the
conversation history. \cite{xu-hu-2018-end} propose using a pointer
network \cite{NIPS2015_5866} for extracting slot values from the
history. More recently, hybrid approaches which combine the candidate-set
and slot-value generation approaches have appeared~\cite{goel2019hyst,trade2019}. 

Our work is most similar to TRADE~\cite{trade2019}, and extends it by proposing
self~\cite{cheng-etal-2016-long} and
cross-attention \cite{BahdanauCB14} mechanism for
capturing slot and history correlations. Attention based archirectures like the
Transformer \cite{Vaswani:2017} and architectures that extend it, like
BERT \cite{devlin-etal-2019-bert} and RoBERTa \cite{roberta}, achieve the
current state-of-the-arts for many NLP tasks. We are also inspired by the work in
reading comprehension where cross attention is used to compute relations between
a long passage and a query question \cite{zhu2018sdnet,Chen_2017}.

For benchmarking, DSTC challenges provide a popular experimentation framework
and dialog data collected through human-machine interactions. Initially, they
focused on single domain systems like bus routes \cite{WilliamsRRB13}. Wizard-of-Oz (WOZ) is also a
popular framework used to collect human-human dialogs that reflect the target
human-machine behavior \cite{wen-etal-2017-network,asri2017frames}. Recently,
the MultiWOZ 2.0 dataset, collected through WOZ for multiple domains, was
introduced to address the lack of a large multi-domain DST
benchmark \cite{budzianowski-etal-2018-multiwoz}. \cite{ericMultiWOZ21} released
an updated version, called MultiWOZ 2.1, which contains annotation corrections
and new benchmark results using the current state-of-the-art approaches. Here, we use the MultiWOZ 2.1
dataset as our benchmark.

\section{Model Architecture}
\begin{figure*}
	\centering
	\includegraphics[width=15cm,height=9cm]{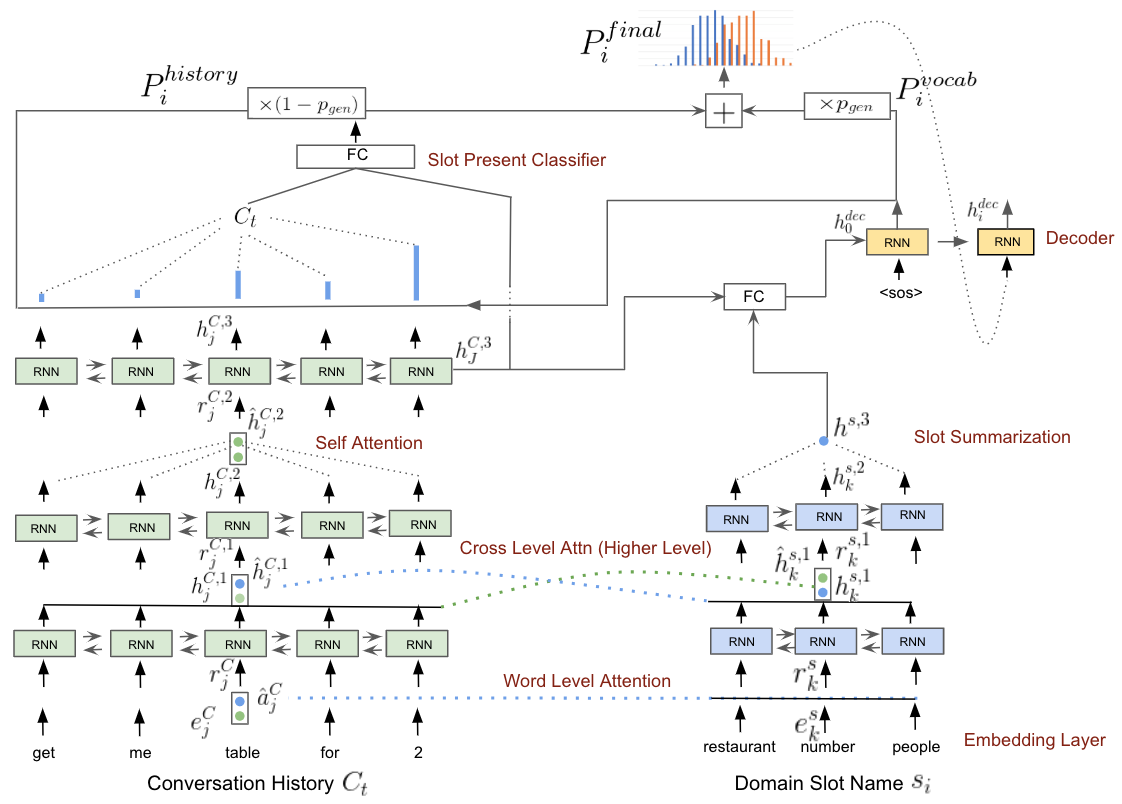}
	\caption{Model Architecture}
	\label{ModelArch}
\end{figure*}
\subsection{Problem Statement}
\label{problem_statement}
Let's denote conversation history till turn $t$ as $C_t = \{U_1, A_1, U_2, A_2, ... U_t\}$, where $U_i$ and $A_i$ represents the user's utterance and agent's response at the $i^{th}$ turn. Let $S = \{s_1, s_2, ... , s_n\}$ denote the set of all $n$ possible slots across all domains. Let $DST_t = \{s_1:v_1, s_2:v_2, ... ,s_n:v_n\}$ denote the dialog state at turn $t$, which contains all slots $s_i$ and their corresponding values $v_i$. Slots that are not mentioned in the dialog history take a $``none"$ value. DST consists of predicting slot values for all slots $s_i$ at each turn $t$, given the conversation history $C_t$.



\subsection{Model Architecture Overview}
\label{architecture}

Our model encodes both the slot name $s_i$ and the
conversation history so far $C_t$, and then decodes the slot value $v_i$,
outputting words or special symbols for $``none"$ and $``dontcare"$ values. Our
proposed model consists of an encoder $\text{Enc}_{slot}$ for the slot name, an
encoder $\text{Enc}_{conv}$ for the conversation history, a decoder
$\text{Dec}_{gen}$ that generates the slot value, and a three-class ``slot gate"
classifier $SG$ that predicts special symbols $\{none, dontcare, gen\}$, which
will be described in detail later on. The model weights are shared between the
slots, which makes the model more robust and scalable.

This architecture is similar to \cite{trade2019}. We propose
modifications to the encoders in order to capture more fine grained dependencies
between the slot name and the conversation history. Also, note the domain and
slot names are concatenated into a single slot description, which we refer to as
slot name for simplicity, and encoded via the slot encoder
$\text{Enc}_{slot}$. Figure \ref{ModelArch} illustrates the proposed
architecture which we refer to as Multi-Attention DST (MA-DST).


\subsection{Encoders}
\label{encoders}

Our proposed slot $s_i$ and conversation history $C_t$ encoders use three stages of attention, specifically low-level cross-attention on the words, higher level cross-attention on the hidden state representations, and self-attention within the dialog history. Below we describe the encoders bottom-up.

\subsubsection{Enriched Word Embedding Layer}
For both $C_t$ and $s_i$, we first project each word into a low-dimensional space. We use a 300-dimensional GloVE embedding  \cite{glove}, and a 100-dimensional character embedding, both of which gets fine-tuned. For the conversation history $C_t$, we also add a 5-dimensional POS tag embedding and a 5-dimensional NER tag embeddings. We also use the turn index for each word as a feature and initialize it as a 5-dimensional embedding. 

To capture the contextual meaning of words, we additionally use contextual ELMo embeddings \cite{Peters_2018}. We compute 1024-dimensional ELMo embeddings for both $C_t$ and $s_i$ by taking a weighted average of the different ELMo layers' outputs. Instead of fine-tuning parameters of all the ELMo layers, we just learn these combination weights while training the model.  All the word-level embeddings are concatenated to generate an enriched, contextual word-level representation $e$.
\begin{multline}
e = [\text{GloVE}(w), \text{CharEmbedding}(w), \text{ELMo}(w), \\\text{POS-tag}(w), \text{NER-tag}(w), \text{position-tag}(w)] 
\end{multline}

\subsubsection{Word-Level Cross-Attention Layer}
To highlight the words in the conversation history $C_t$ relevant to the slot $s_i$, we add a word-to-word attention from conversation history to the slot. For computing the attention weights, we used symmetric scaled multiplicative attention \cite{huang2017fusionnet} with a ReLU non-linearity. The weights are calculated according to equation \ref{eq1} and used according to equation \ref{eq2} to obtain the attended vector for each word in the conversation. 
\begin{align}
\alpha_{jk} &=  \frac{\exp(f(We_{j}^C)Df(We_{k}^s))}{\sum_{k=1}^K\exp(f(We_{j}^C)Df(We_{k}^s))} \label{eq1} \\ 
\hat{a}_{j}^C &= \sum_{k=1}^{K} \alpha_{jk}*e_{k}^s \label{eq2} 
\end{align}

Here, $e_{j}^C$ and $e_{k}^s$ correspond to the word embedding of the $j^{th}$ word in the conversation and $k^{th}$ word in the slot. The length of the slot is denoted by $K$. $f$ denotes a non-linear activation, which here is a ReLU. To get the representation $r_j$ for each word in the conversation history, we concatenate the attended vector with the initial word embedding:
.    \begin{align}
r_{j}^C &= [e_{j}^C, \hat{a}_{j}^C]
\end{align}
For the slot representation for each word $k$ in the slot name, we use the word embedding $r_{k}^s = e_{k}^s$.

Note that symmetric scaled multiplicative attention with ReLU non-linearity is used in all attention computations of our proposed models, as we empirically found that it gives better performance compared to other attention variants.(e.g. multiplicative, scaled multiplicative, additive).

\subsubsection{First Layer RNN}
The computed representations $r_{k}^s$ and $r_{j}^C$ for each word in the slot name and the conversation history respectively, are then passed through a Gated Recurrent Unit (GRU) \cite{chung2014empirical} in order to model the temporal interactions between the words and get a contextual representation. For each of $C_t$ and $s_i$, we use bidirectional GRUs and obtain the hidden contextual representation by averaging the hidden states of each GRU direction per time step:
\begin{align}
H_{1}^{C} &= [h_{1}^{C,1}, h_{2}^{C,1}, ..., h_{J}^{C,1}]  &=  GRU([r_{1}^{C}, r_{2}^C, ...., r_{J}^C])\\
H_{1}^{s} &= [h_{1}^{s,1}, h_{2}^{s,1}, ..., h_{K}^{s,1}]  &=  GRU([r_{1}^s, r_{2}^s, ...., r_{K}^s])
\end{align}
Here, $H_{1}^{C}$ and $H_{1}^{s}$ are the sequences of encoded representations for the conversational history and slot name respectively, output by the first bidirectional GRU layer (assuming K is the slot name length and J the conversation history length in number of words). 


\subsubsection{Higher Level Cross Attention Layer}

We add a cross-attention network on top of the the base RNN layer to attend over higher level representations generated by the previous RNN layer, i.e. $H_{1}^{C}$ and $H_{1}^{s}$. We used two-way cross-attention network, one from conversation history ($H_{1}^{C}$) to the slot ($H_{1}^{s}$) and the other in the opposite direction. This is inspired by several works in reading comprehension where cross attention is used to compute relations between a long passage and a query question \cite{Weissenborn_2017,Chen_2017}. 

The \textbf{Slot to Conversation History attention sub-network} helps in highlighting the words in the conversation which are relevant to the slot for which we want to generate the value. Similar to the word level attention, the attention weights are calculated by equation \ref{eq3}.
\begin{align}
\alpha_{jk} &=  \frac{\exp(f(Vh_j^{C,1})D'f(Vh_k^{s,1}))}{\sum_{j=1}^J\exp(f(Vh_j^{C,1})D'f(Vh_k^{s,1}))} \label{eq3} \\ 
\hat{h}_{k}^{s,1} &= \sum_{j=1}^J \alpha_{jk}*h_{j}^{C,1} \label{eq4}  
\end{align}
We fuse the attention vector $\hat{h}_k^{s,1}$ with it's corresponding hidden state $h_k^{s,1}$ for each word in the slot name as follows:
\begin{align}
r_k^{s,1} = &[h_k^{s,1}, \hat{h}_k^{s,1},  \hat{h}_k^{s,1} + h_k^{s,1}, \hat{h}_k^{s,1}*h_k^{s,1}]   \label{eq5}
\end{align}
where, $*$ is the element wise dot product operation. 


Similarly, the \textbf{Conversation to Slot attention sub-network} computes attention weights to highlight which words in the slot name are most relevant to each word in the conversation history. This enriches the word representation in the conversation history $h_j^{C,1}$ with an attention based representation $\hat{h}_j^{C,1}$, resulting in a new representation  $r_j^{C,1}$. All computations are similar as in the Slot to Conversation History attention, but in the reverse direction.

\subsubsection{Second Layer RNN}
The representations $r_k^{s,1}$ and $r_j^{C,1}$ are then passed through a second bidirectional GRU layer, to obtain $h_k^{s,2}$ and $h_j^{C,2}$. This helps in fusing these vectors together along with the temporal information.

\subsubsection{Self Attention Layer}
We add a self attention network on top of the conversation representation $h_j^{C,2}$. This layer helps resolve correlation between words across utterances in the conversation history. We introduce this sub-network to address cases where the user refers to slot values that are present in previous utterances, which is a common phenomenon in dialogs, especially multi-domain ones. Self attention is computed as:
\begin{align}
\alpha_{ji} &=  \frac{\exp(f(Wh_j^{C,2})Df(Wh_i^{C,2}))}{\sum_{i=1}^J\exp(f(Wh_j^{C,2})Df(Wh_i^{C,2}))} \label{eq6} \\ 
\hat{h}_{j}^{C,2} &=  \sum_{i=1}^J \alpha_{ji}*h_{i}^{C,2} \label{eq7} 
\end{align}

The final representation $r_j^{C,2}$ for each word in the conversation is the merged representation of self-attended vector $\hat{h}_{j}^{C,2}$ and the hidden state $h_{j}^{C,2}$, merged according to equation \ref{eq5}. 

\subsubsection{Third Layer RNN and Slot Summarization} We use a third layer RNN to get the final representation for the conversation history
\begin{align}
h_j^{C,3} =  GRU (r_j^{C,2}), j=1..J \label{eq8}
\end{align}

Since the slot name  is much shorter in length than the conversation history, it can be encoded with less information. Instead of using an additional RNN, we summarize the slot using a linear transformation to reduce the slot representation into a single vector.
\begin{align}
\alpha_k &= w^\intercal*h_k^{s,2}\\ 
h^{s,3} &= \sum_{k=1}^K \alpha_k*h_k^{s,2}
\end{align}
where, $w^\intercal$ is the parameter which is learnt during training. 

Finally, $H^{C,3} = [h_1^{C,3}, h_2^{C,3}, ..., h_J^{C,3}]$ is the per word representation for the conversation history, while $h^{s,3}$ is the summarized slot name representation, both of which will be used at the decoding step.

\subsection{Decoder and Slot Gate classifier}
\label{decoder}


The decoder network is a GRU that decodes the value $v_i$ for slot $s_i$. At
each decoding step $i$ that computes each word in the slot value, the network
computes two distributions: a distribution over all in-vocabulary words (word
generation distribution) and one over all words in the conversation history
(word history distribution). This allows the decoder to generate unseen words
that appear in the conversation history but are not present in the vocabulary of
the training data. This formulation removes the dependency of having a
predefined ontology that contains all the possible slot values, which is
restrictive for free-form slots. Because of the ability to generate unseen slot values, the network is well-suited for zero-shot use cases.

We initialize the decoder by combining the last hidden state of the conversation history representation and the summarized slot representation: 
\begin{align}
h_0^{dec} = W[h_J^{C,3}, h^{s,3}]
\end{align}
where W is a learnable parameter. At each decoding time-step $i$, the decoder generates a probability distribution over the vocabulary: 
\begin{align}
P_i^{vocab} &= Softmax(W*h_i^{dec}) 
\end{align}
The decoder also generates a probability distribution over words in the conversation history $P_{history}$ by using a pointer network \cite{See_2017}, i.e., computing attention weights for each word in the conversation history.

To generate the final vocabulary distribution, we take a weighted sum of $P_{history}$ and $P_{vocab}$:     \begin{align}
P_i^{final} &= p_i^{gen}*P_i^{vocab} + (1-p_i^{gen})*P_i^{history} 
\end{align}
Where $p_{gen}$ is the probability to generate a word as opposed to copy from the history, and is calculated at each decoder time step. 

To avoid running the decoder for slots not present in the conversation, we also train a Slot Gate classifier\cite{trade2019}. This is a 3-way classifier which predicts among the following classes $\{none, dontcare, gen\}$. Only when the classifier predicts $gen$ we decode the slot value. When the classifier predicts $``none''$ we assume that the slot is not present and takes a $``none''$ value in the state, and when it predicts $``dontcare"$, we assume the user does not care about the slot value (this appears commonly in dialog and therefore $``dontcare"$ is a special value for DST systems).

The network is trained in a multi-task manner using standard cross entropy loss. We combine the losses of the slot generator (decoder) and the SG classifier as follows:
\begin{align}
Loss_{combined} &= Loss_{generator} + \gamma*Loss_{classifier} 
\label{eq9}
\end{align}
where $\gamma$ is a hyperparameter that is optimized empirically.

\section{Dataset}
We evaluate our approach on MultiWOZ, a multi-domain Wizard-of-Oz dataset. MultiWOZ 2.0 is a recent dataset of labeled human-human written conversations spanning multiple domains and topics \cite{budzianowski-etal-2018-multiwoz}. As of now, it is the largest labeled, goal-oriented, human-human conversational dataset with around 10k dialogs, each with an average of 13.67 turns. The data spans seven domains and 37 slot types. Due to patterns of annotation errors found in MultiWOZ 2.0, \cite{ericMultiWOZ21} re-annotated the data and released a MultiWOZ 2.1 version, which corrected a significant number of errors. Table \ref{newDataStats} mentions the percentage of slots in each domain whose values changed with the MultiWOZ 2.1 re-annotation. 

For all our experiments, we use MultiWOZ 2.1 data, which is shown to be cleaner and more challenging because many slots are now correctly annotated with their corrected values or $dontcare$ instead of $none$. We are using only five domains out of the available seven - namely $(\text{restaurant}, \text{hotel}, \text{attraction}, \text{taxi}, \text{train})$ - since the other two domains $(\text{bus}, \text{police})$ are only present in the training set. We use the provided train/dev/test split for our experiments. 
    
\begin{table}[h!]
\centering
\begin{tabular}{p{1cm} p{1.5cm}  p{1cm} p{1cm} p{1cm} p{1cm}}
 \multicolumn{6}{c}{Slot Values Updated in MultiWOZ 2.1} \\
   & Restaurant & Taxi & Hotel & Train & Attraction\\
 \hline
 Train  & 13.64 & 3.65 & 26.89 & 7.04  & 12.69\\
 Dev    & 22.04 & 3.18 & 20.93 & 5.88  & 12.82\\
 Test   & 19.33 & 3.95 & 24.70 & 10.59 & 16.12\\
  \hline
 \end{tabular}
 \caption{Percentage of slot values that changed in MultiWOZ 2.1 compared to MultiWOZ 2.0. }
 \label{newDataStats}
\end{table}

\section{Evaluation}
In this section we first describe the evaluation metrics and then present the results of our experiments.  
\subsection{Metrics}
Following are the metrics used to evaluate DST models:
\begin{itemize}
    \item \textbf{Average Slot Accuracy}: The average slot accuracy is defined as the fraction of slots for which the model predicts the correct slot value. For an individual dialog turn $D_{t}$, the average slot accuracy is defined as follows: 
    \begin{align}
        \sum_{i=1}^{n}  \mathbbm{1_{y_{i} = \hat{y}_{i}}}
    \end{align}
    where $y_{i}$ and  $\hat{y}_{i}$ are ground truth and predicted slot value for $s_{i}$ respectively, $n$ is the total number of slots, and $\mathbbm{1_{x=y}}$ is an indicator variable that is 1 if and only if $x=y$.
    \item \textbf{Joint Goal Accuracy}: The joint goal accuracy is defined as the fraction of dialog turns for which the values $v_i$ for all slots $s_i$ are predicted correctly. If we have $n$ total slots we want to track, the joint goal accuracy for an individual dialog turn $D_{t}$ is defined as follows:      
    \begin{align}
        \mathbbm{1_{({(\sum_{i=1}^{n}  \mathbbm{1_{y_{i} = \hat{y}_{i}}}) = n})}}
    \end{align}
\end{itemize}
\subsection{Experiment Details}
We train the encoders to jointly optimize the losses of the slot gate classifier
and the slot value generator decoder. The parameters of the model are shared for
all $(domain, slot)$ pairs, which makes this model scalable to a large number of
domain and slots. We train the model using  stochastic gradient
descent  and use the Adam Optimizer. We empirically optimized the learning
rate in the range $[0.0005 - 0.001]$ and used $0.0005$ for the final model,
while we kept betas as $(0.9, 0.999)$ and epsilon 1x$10^{-08}$. We used a batch size of
four dialog turns and for each turn we generate all 30 slot values. We decayed
the learning rate after regular intervals ($3$ epochs) by a factor of $\theta$
($0.25$), which was empirically optimized. For ELMo, we kept a dropout of 0.5 for
the contexual embedding and used $l_{2}$  regularization for the weights of
ELMo. We used a dropout of 0.2 for all the layers everywhere else. For word
embeddings, we used 300-dimensional GloVe embeddings and 100-dimensional
character embeddings. For all the GRU and attention layers the 
hidden size is kept at 400. The weight $\gamma$ for the multi-task loss function in equation \ref{eq9} is kept at $1$.


\subsection{Results}
In this section, we present the results for our model. We measure the quality of the model on joint goal accuracy and average slot accuracy, as described earlier. As our baseline for comparison, we consider the TRADE model \cite{trade2019}, which is the present state of the art for MultiWOZ. To have a fair comparison, we report the numbers on the corrected MultiWOZ 2.1 dataset for both models.

In Table \ref{singleDomainResult}, we present the results for DST on single-domain data. We create the train, dev, and test splits of the data for a particular domain by filtering for dialogs which only contain that domain. As shown in table \ref{singleDomainResult}, MA-DST outperforms TRADE for all five domains, improving the joint goal accuracy by up to 7\% absolute as well as the average slot accuracy by up to 5\% absolute.
\begin{table}[h!]
\centering
\begin{tabular}{ p{2cm} p{1cm} p{1cm} | p{1cm} p{1cm} }
 \hline
 \multicolumn{5}{c}{Single Domain} \\
 \hline
 \multicolumn{1}{c}{ } &\multicolumn{2}{c|}{MA-DST} &  \multicolumn{2}{c}{TRADE\footnotetext{working}} \\
 \hline
 Domain & Joint & Slot & Joint & Slot\\
 \hline
 Hotel   & {57.70} & {93.41} & 50.25 & 90.48\\
 Train   & {76.47} & {94.87} & 74.47  & 94.30 \\
 Taxi    & {76.55} & {91.25} & 70.18 & 86.27\\
 Restaurant & {66.33} & {93.86} & 66.02 & 93.73\\
 Attraction & {72.49} & {89.38} & 68.48 & 86.89\\
 \hline
 \end{tabular}
 \caption{Joint goal and slot accuracy of MA-DST and TRADE on 5 single-domain datasets from MultiWOZ 2.1 }
 \label{singleDomainResult}
\end{table}

Table \ref{multiDomainResult} shows results for the multi-domain setting, where we combine all available domains during training and evaluation. We compare the the accuracy of MA-DST with the TRADE baseline and four additional ablation variants of our model. These four variants capture the contribution of the different sub-networks and layers in MA-DST on top of the base encoder-decoder architecture, which is called ``Our Base Model" in Table \ref{multiDomainResult}. Our full proposed MA-DST model achieves the highest performance on joint goal accuracy and average slot accuracy, surpassing the current state-of-the-art performance. Each of the additional layers of self and cross-attention contribute to progressively higher accuracy for both metrics.

\begin{table}[h!]
\centering
 \begin{tabular}{p{6cm}| c c} 
 \hline
 \multicolumn{3}{c}{Multi Domain} \\
 \hline
 Model & Joint & Slot \\ [0.5ex] 
 \hline
 Baseline (TRADE~\cite{trade2019}) & 45.6 & 96.62 \\ 
 Our Base Model & 44.0 & 96.15 \\
 + Slot Gate
 + Word-Level Cross-Attention &  {47.60}  & {97.01}\\
 + Higher-Level Cross-Attention & {49.56}  & {97.15}\\
 + Self-Attention + Slot Summarizer & {50.55}  & {97.21}\\
 + ELMo (MA-DST) & 51.04  & 97.28\\
 + Ensemble & 51.88 & 97.39 \\
 \hline
 \end{tabular}
 \caption{Joint goal and slot accuracy of different models in the all-domain setting of MultiWOZ 2.1}
  \label{multiDomainResult}
\end{table}

\begin{table}[h!]
\centering
\begin{tabular}{ p{2cm} p{1cm} | p{1cm} }
 \hline
 \multicolumn{3}{c}{Zero Shot Experiment} \\
 \hline
 \multicolumn{1}{c}{ } &\multicolumn{1}{c|}{MA-DST} &  \multicolumn{1}{c}{TRADE} \\
 \hline
 Domain & Joint  & Joint \\
 \hline
 Hotel   & {16.28} & 14.20 \\
 Train   & {22.76} & 22.39  \\
 Taxi    & {59.27} & 59.21 \\
 Restaurant & {13.56} & 12.59 \\
 Attraction & {22.46} & 20.06\\
 \hline
 \end{tabular}
 \caption{Joint goal accuracy of MA-DST and TRADE in the zero shot setting for the five domains of MultiWOZ 2.1.}
 \label{zeroShotResult}
\end{table}

In Table \ref{zeroShotResult} we present the zero shot results. For these experiments, the test set contains only dialogs from the target domain while the training set contains only dialogs from the other four domains. As shown in Table \ref{zeroShotResult}, MA-DST outperforms TRADE's state-of-the-art result by up to 2\% on the joint goal accuracy metric.

\begin{figure}
	\centering
    \includegraphics[width=0.5\textwidth]{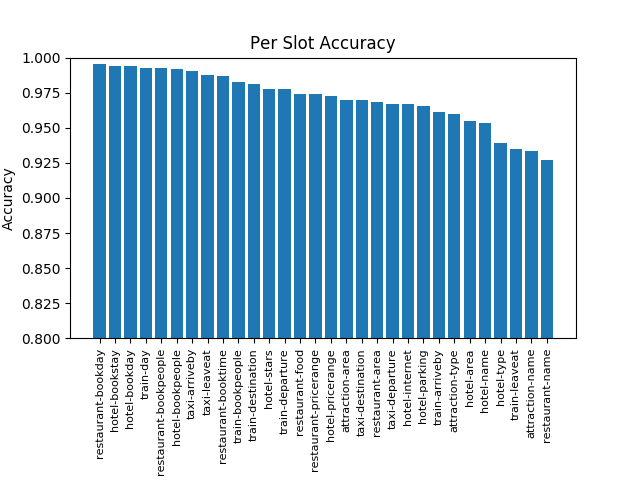}
    \caption{Per Slot Accuracy on test set in multi-domain setting for MA-DST.}
    \label{slotAcc}
\end{figure}

\subsection{Error Analysis}
\begin{table}[h!]
    \centering
    \begin{tabular}{ p{2cm} | p{1.5cm} | p{1.5cm} }
     \multicolumn{3}{c}{Domain Level Statistics} \\
     \hline
     Domain & F1-Score  & Slot Acc. \\
     \hline
     Hotel   & {0.90} & 97.11 \\
     Train   & {0.92} & 97.16  \\
     Taxi    & {0.71} & 97.87 \\
     Restaurant & {0.94} & 97.41 \\
     Attraction & {0.87} & 95.46\\
     \hline
     \end{tabular}
     \caption{F1 Score and Average Slot Accuracy Domain Wise}
     \label{domainLvlStats}
    \end{table}
    
In this section we analyze the errors being made by the model on MultiWoz 2.1 dataset. Table \ref{domainLvlStats} shows the Average Slot Accuracy and F1-Score for each domain. In terms of F1-Score, the model performs worse for Taxi domain. The average slot accuracy for Taxi domain is high because a vast number of taxi domain's slots are $``none"$ (i.e. not present in the dialog), which model easily identifies. Figure \ref{slotAcc} shows the per-slot accuracy in the all-domain setting, in descending order of performance. As seen from Figure \ref{slotAcc}, the MA-DST model tends to make the most errors for open-ended slots such as $\text{restaurant-name}$, $\text{attraction-name}$, $\text{hotel-name}$, $\text{train-leaveat}$. These slots are difficult to predict for the model because, unlike categorical slots, these slots can take on a large number of possible values and are more likely to encounter unseen values. On the other end of spectrum, we have slots like $\text{restaurant-bookday}$, 
$\text{hotel-bookstay}$,
$\text{hotel-bookday}$, and
$\text{train-day}$, 
for which the model is able to achieve more than $99\%$ in terms of average slot accuracy. As expected, most of the top-perfoming slots are categorical, i.e. they can take only a small number of different values from a pre-defined set. 

\begin{figure}
	\centering
	\includegraphics[width=0.5\textwidth]{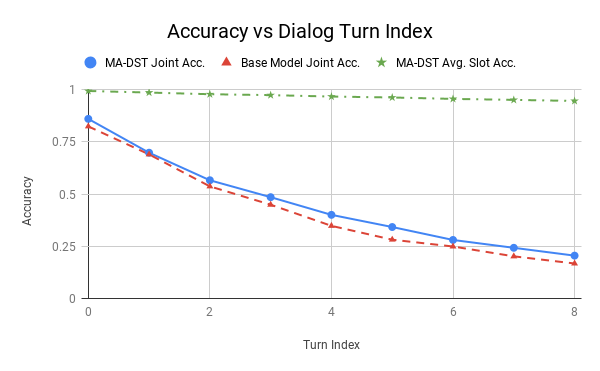}
	\caption{Accuracy of MA-DST and our base model aggregated by dialog turns.}
	\label{TurnStats}
\end{figure}

Figure \ref{TurnStats} analyzes the relationship between depth of conversation and accuracy of MA-DST. To calculate this, we first bucket the dialog turns according to their turn index, and calculate the joint goal accuracy and average slot accuracy for each bucket. As shown in Figure \ref{TurnStats}, the joint goal accuracy and average slot accuracy for MA-DST is around $88\%$ and $99\%$ for turn 0, and it decreases to $8\%$ and $92\%$ for turn 10. As expected, we can see that the model's performance degrades as the conversation becomes longer. This can be explained by the fact that longer conversations tend to be more complex and can have long-range dependencies. To study the effect of attention layers, we compare the joint goal accuracy of our base model, which does not have the attention layers, and MA-DST for each turn. As can be seen from Figure \ref{TurnStats}, MA-DST performs better than our base model, which doesn't have the additional attention layers, for both earlier and later turns by an average margin of $4\%$. 

To further analyze what type of errors the model is making, we manually analyzed the model's output for 20 randomly selected dialogs. Around $36\%$ of the errors are because of wrong annotations, i.e., the model predicted the slot value correctly but the target label was wrong. For e.g. in turn 5 of PMUL3158, $restaurant$ $book$ $time$ is annotated as $none$, while user has mentioned 17:45 as the booking time. These kind of annotation errors are unavoidable. The other common error we observed was of model getting confused among slots of same types. For e.g. in turn 3 of dialog PMUL4547, model populates $attraction$ $name$ and $hotel$ $name$ with \text{``The Junction"}, as user didn't specify in the utterance whether \text{``The Junction"} is attraction or a hotel. Because of similar reason, we also see model confusing between $taxi$ $destination$ and $taxi$ $departure$ slot quite a number of times.  The other common type of error model makes is by generating slot value which varies from the ground-truth by a word or character. For e.g. for dialog MUL2432, the model generates the value of $restaurant$ $book$ $time$ as 15.15 by directly copying it from the user utterance, however, the label is 15:15 according to the ontology. This kind of error can be solved by fuzzy match between ontology and model's prediction, but it will introduce dependency on the ontology. We also observed that model's accuracy for slot values which were \text{``dontcare''} was only $60\%$. We also observed that there are lots of annotation errors for slots with \text{``dontcare''} in the training set, thus making it difficult for the model to learn.


\section{Conclusion}
We propose a new architecture for dialog state tracking that uses multiple
levels of attention to better encode relationships between the conversation
history and slot semantics and resolve long-range cross-domain
coreferences. Like TRADE \cite{trade2019}, it does not rely on knowing a complete list of
possible values for a slot beforehand and both generate values from the
vocabulary and copy values from the conversation history. It also shares the
same model weights for all $(domain, slot)$ pairs so it can easily be adapted to
new domains and applied in a zero-shot or few-shot setting. We achieve new
state-of-the-art joint goal accuracy on the updated MultiWOZ 2.1 dataset of
51\%.  In the zero-shot setting we improve the state-of-the-art by over
2\%. In the future, it is worth exploring whether the state can be carried from
the previous turn to predict the state for the current turn (rather than
starting from scratch for each turn). Finally, it may be useful to capture dependencies or correlations
between slots rather than independently generating values for each one of them.

\bibliography{dst_refs}
\bibliographystyle{aaai}
\end{document}